# An Improved ACS Algorithm for the Solutions of Larger TSP Problems

**Md. Rakib Hassan, Md. Kamrul Hasan and M.M.A Hashem**

Department of Computer Science & Engineering
Khulna University of Engineering & Technology,
Khulna-9203, Bangladesh

## ABSTRACT

**Solving large traveling salesman problem (TSP) in an efficient way is a challenging area for the researchers of computer science. This paper presents a modified version of the ant colony system (ACS) algorithm called Red-Black Ant Colony System (RB-ACS) for the solutions of TSP which is the most prominent member of the combinatorial optimization problem. RB-ACS uses the concept of ant colony system together with the parallel search of genetic algorithm for obtaining the optimal solutions quickly. In this paper, it is shown that the proposed RB-ACS algorithm yields significantly better performance than the existing best-known algorithms.**

## 1. INTRODUCTION

To experiment with the modified ant colony system, *traveling salesman problem (TSP)* [1], [2] is chosen because it exhibits all the aspects of combinatorial optimization problems and serves as the benchmark problem for evolutionary methods, neural networks, genetic algorithm, simulated annealing, greedy algorithm and the well-known ant colony system *(ACS)* [3]-[6]. In TSP, a number of cities are given which are connected with each other by edges. The connecting edges have some finite distances. A salesman starts his tour from any city and after traveling each city only once, he returns to the starting city again. Solution of this problem helps the salesman by finding the shortest tour of visiting each city exactly once.

The existing algorithms for solving the traveling salesman problem are not efficient enough. When the number of cities increases, the time or space required solving the problem increases exponentially. The ant colony system algorithm solves the problem in a reasonable time and produces optimal solutions. But when the dimension of the problem increases, it is time-consuming to produce the result and the result is not often optimal. So, a modification was required for solving large traveling salesman problems. The purpose of the modification or improvement is to solve the problem in a short time and at the same time, the result is needed to be optimized. This problem inspired us for a new algorithm by improving the ant colony system.

In this algorithm, two groups of artificial ants are used which cooperate with other ants to find the solution of a given problem by exchanging information via pheromone deposited on graph edges [3]. Pheromone is a chemical which is deposited on the ground by ants while walking. Ants prefer the paths where more pheromone is deposited. This new modification of the *ACS* is experimented using some of the benchmark problems of the traveling salesman problem and it is observed that the proposed *RB-ACS* yields better results than other well-known algorithms.

## 2. ANT COLONY SYSTEM (ACS)

In ACS [3], a number of artificial ants are used which are initially placed randomly in the cities. Each ant builds a tour, that is, a feasible solution to the TSP by repeatedly applying a stochastic greedy rule which is called here the *state transition rule* [3], [4]. The state transition rule provides a direct way to balance between exploration of new edges and exploitation of the accumulated knowledge about the problem. When an ant is positioned on city *r*, then it chooses the next city s by applying the following state transition rule:

$$s = \begin{cases} \arg\max_{u \in J_k(r)} \{[\tau(r,u)] \cdot [\eta(r,u)]^{\beta}\} & \text{if } q \le q_0 \quad (\text{exploitation}) \\ S & \text{otherwise} \quad (\text{biased exploration}) \end{cases} \tag{1}$$

Here, $q$ is a random number uniformly distributed in [0 .. 1], $q_0$ is a parameter ($0 \le q_0 \le 1$), $\beta$ is the parameter that determines the relative importance between pheromone and distance, $\tau(r,u)$ is the amount of pheromone between the nodes r and u, $\eta = 1/\delta$ is the inverse of the distance $\delta(r,s)$, $J_k(r)$ is the set of cities yet to be visited by ant k positioned on city r. $S$ is a random variable selected according to the probability distribution [3], [4] given as follows:

$$p_k(r,s) = \begin{cases} \dfrac{[\tau(r,s)] \cdot [\eta(r,s)]^{\beta}}{\sum_{u \in J_K(r)} [\tau(r,u)] \cdot [\eta(r,u)]^{\beta}} & \text{if } s \in J_k(r) \\ 0 & \text{otherwise} \end{cases} \tag{2}$$

This state transition rule is used for the transitions towards nodes connected by short edges and with a large amount of pheromone. The parameter $q_0$ determines the relative importance of exploitation versus exploration and n is the number of cities of the traveling salesman problem.

While constructing a tour, an ant also modifies the amount of pheromone on the visited edges by applying the *local updating rule* [4]. The local updating rule is defined using the Eqn. (3):

$$\tau(r,s) = (1-\rho)\cdot\tau(r,s) + \rho\cdot\Delta\rho(r,s) \qquad (3)$$

Here, 0< ρ<1 is the local pheromone decay parameter. The effect of local-updating is to make the desirability of edges change dynamically. Every time an ant uses an edge, it becomes slightly less desirable because it loses some of its pheromone making them less desirable for future ants and allowing for the search of new, possibly better tours in the neighborhood of the previous best tour.

Once all ants have terminated their tour, the amount of pheromone on edges is modified again by applying the *global updating rule* [4]. But only the globally best ant, constructing the shortest tour, is allowed to deposit pheromone. This is done due to make the search more directed. The pheromone level is updated by applying the global updating rule of Eqn. (4):

$$\tau(r,s) = (1-\alpha)\cdot\tau(r,s) + \alpha\cdot\Delta\tau(r,s) \qquad (4)$$

$$where \ \ \Delta\tau(r,s) = \begin{cases} (L_{gb})^{-1} & if\ (r,s) \in global-best-tour \\ 0 & otherwise \end{cases} \qquad (5)$$

Here, 0< α<1 is the global pheromone decay parameter [4] and $L_{gb}$ is the length of the globally best tour from the beginning of the trial.

## 3. RB-ACS: THE PROPOSED APPROACH

This paper presents an improvement of the *ant colony system* for solving large traveling salesman problems. The modified approach is named as *Red-Black Ant Colony System.* The *RB- ACS* is compared with the ant colony system [3]-[6], ant colony optimization with multiple ant clans (ACOMAC) [7], multiple nearest neighbor (NN) [7] and dual nearest neighbor (DNN) [7] to ACS and with the NN and DNN to ACOMAC.

### *3.1 The Modifications to the Ant Colony System*

Although the proposed *RB-ACS* uses the basic concept of the ant colony system, it has some major modifications. The changes that are made to the ACS are as follows:

i. *Pheromone Initialization:* In *RB-ACS*, the initial pheromone of the edges is set according to the Eqn. (6). So, the edges which have higher cost, obtain lower pheromone making the desirability of that edge to decrease whereas in ACS the initial pheromone is constant. This modification helps the proposed algorithm to search the solution in a more directed manner.

$$\tau_{init} = \frac{C}{\cos t(r,s)} \qquad (6)$$

Here, C is a constant and cost(r, s) is the distance between the nodes r and s.

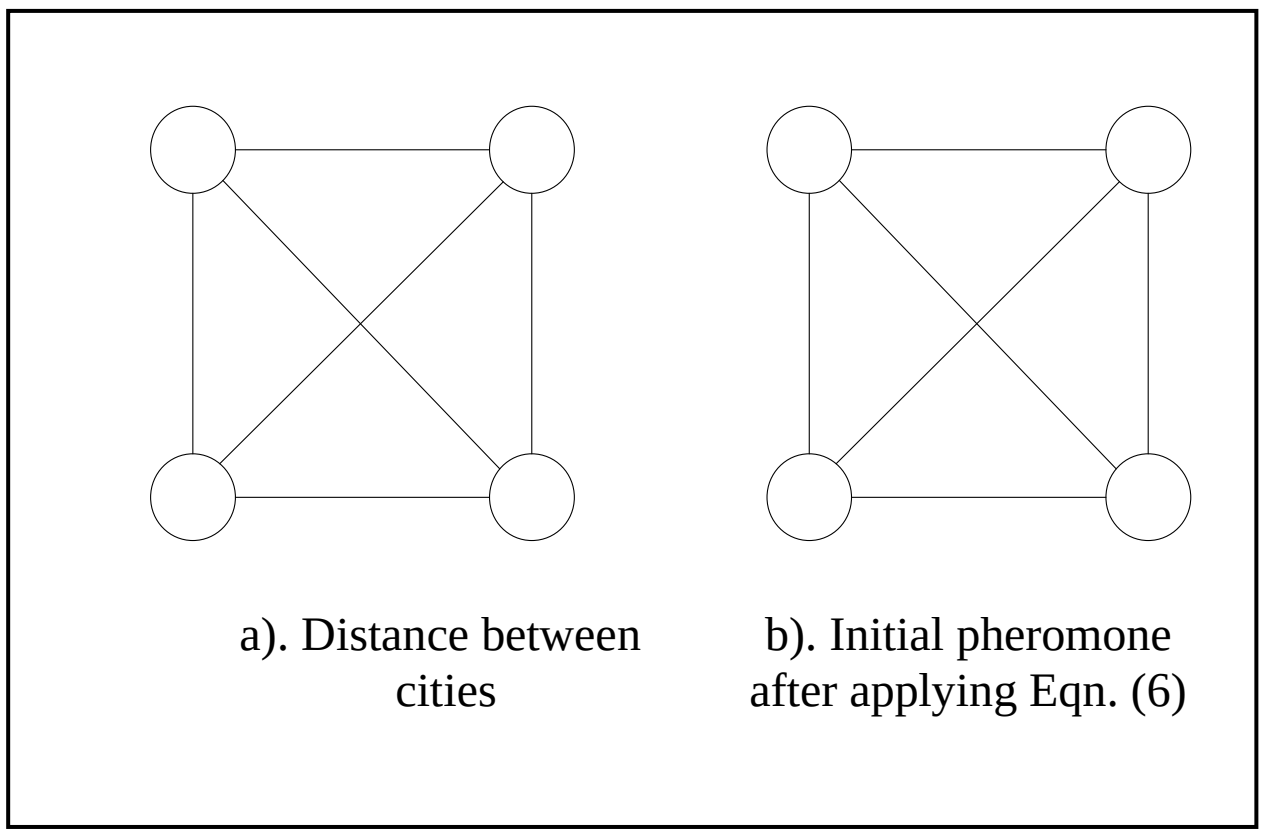

Fig. 1: Application of Eqn. (6) considering C=100

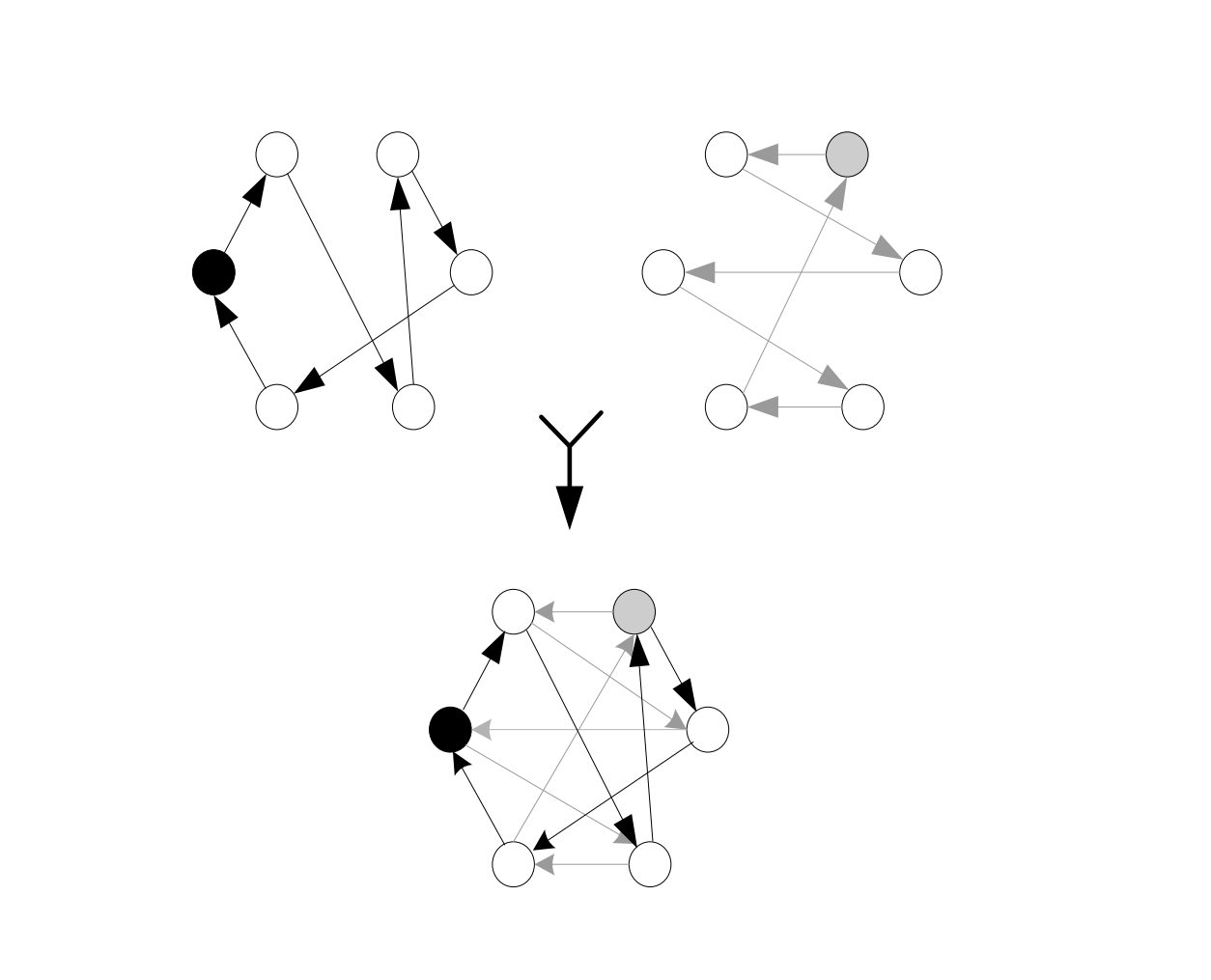

Fig. 2: The local tour of the proposed *RB-ACS* where the black path: 3->1->6->2->4->5->3 and the red path: 2->1->4->3->6->5->2

ii. *Separate local paths:* In ACS, only one group of ants is used to search and the ants may use the path of the other ants. Thus, the search is not fast enough for large TSP problems and the search process may get stuck in the local minima. So, instead of using one group, the proposed approach uses two groups of ants, namely the black group and the other is red group. These two groups search in parallel. They do not follow the path of the other group. Only the ants in the same group may use the same path. As a result, the possibility of getting stuck in local minima is significantly decreased. Fig. 2 shows that the red and black ants use separate paths for solution.

iii. *Different parameter values:* In the proposed *RB-ACS*, the two groups have separate characteristics. They use separate parameter values, especially in the case of local updating rule. This idea is taken from the behavior of real ants. There are different groups of ants each of which has different characteristics. The characteristics that may differ are in the ejaculation of pheromone, evaporation of pheromone, the walking speed of the ants. So, the proposed *RB-ACS* uses separate values for the parameters in local updating and pheromone evaporation.

iv. *Global updating:* When all the ants have made their tours, then global updating rule of Eqn. (4) is applied. In ACS, only the best ant [4] is allowed to deposit pheromone on its path. But in the proposed approach of this thesis, two best ants from each group are allowed to deposit pheromone. This makes the global updating parallel. As a result, the probability of obtaining the optimal solutions increases significantly.

These changes to the ant colony system make the *RB-ACS* to search in a more diversified manner and obtain the optimal or nearly optimal solutions quickly.

### *3.2. Parameter Settings of the Proposed RB-ACS*

The parameters [2] considered here are:

- $\alpha$: pheromone decay parameter [4], $0< \alpha<1$
- $\beta$: relative importance of pheromone versus distance parameter [4], $\beta>0$
- $\rho$: trail persistence, $0\leq\rho<1$ ($1-\rho$ can be interpreted as trail evaporation [2]);
- m: the number of ants, m=20
- C: a constant related to the initial pheromone

$q_0$=0.9, C=100, $\tau_0 = (n.L_{nn})^{-1}$, where $L_{nn}$ is the tour length produced by the nearest neighbor heuristic [12] and $n$ is the number of cities. These values were obtained by a preliminary optimization phase [3]-[5], in which it is found that the experimental optimal values of the parameters were largely independent of the problem, except for $\tau_0$ [4] for where $\tau_0 = (n.L_{nn})^{-1}$.

### *3.3 RB-ACS Algorithm*

Step 1**:** Initialize pheromone and place the ants of each group randomly in each city
Step 2**:** For each city do
    For each ant of the two groups do
        Choose the next city according to the state transition rule
    if(every city is visited) then
        place the ants in their starting cities respectively
    Update the visited edges using local updating rule:
**Step 3:** For each group of ants do
        compute the best paths of each group
    Update the paths
**Step 4:** If (End_condition = True) then
        Print shortest path
    else
        goto Step 2

## 4. EXPERIMENTAL RESULTS

The proposed *RB-ACS* has been verified using different TSP benchmark problems which were obtained in the TSPLIB [9]. The TSPLIB contains sample problems on symmetric and asymmetric traveling salesman problem.

The proposed algorithm was run on three problems: Eil51, Eil76 and Kroa100. And the results were averaged over 30 trials. These results are compared with the ant colony system (ACS), ant colony optimization with multiple ant clans (ACOMAC) [7] using nearest neighbor (NN) heuristic and dual nearest neighbor (DNN) heuristic.

The following figures show the comparison with *RB-ACS*, ACS+DNN and ACOMAC+DNN [7]. According to the simulation results, it is observed that the proposed modification yields desired results in a small number of iterations.

Fig. 3 shows the result using the Eil51 problem with 51 cities of the TSPLIB. It is observed from the figure that the proposed *RB-ACS* is more convergent than other approaches. For Eil51 problem, the optimum length is 426. The *RB-ACS* yields this optimum solution in a small number of iterations and the average length of the problem is 427.5 using this algorithm which is better than other algorithms.

The optimum tour length of Eil76 problem is 538. The average length of the problem produced by the *RB-ACS* is 549.333 which is shown in Fig. 4.

Fig. 5 shows the result of the problem Kroa100 of 100 cities and Table 1 shows the comparison of the *RB-ACS* with ACS (Including ACS+DNN) and ACOMAC (Including ACOMAC+DNN) using three different TSP benchmark problems. The tour length given in this table is of average length.

It is observed from the figures that all the results produced by the proposed approach are better than the other algorithms. The figures are shown on the next page.

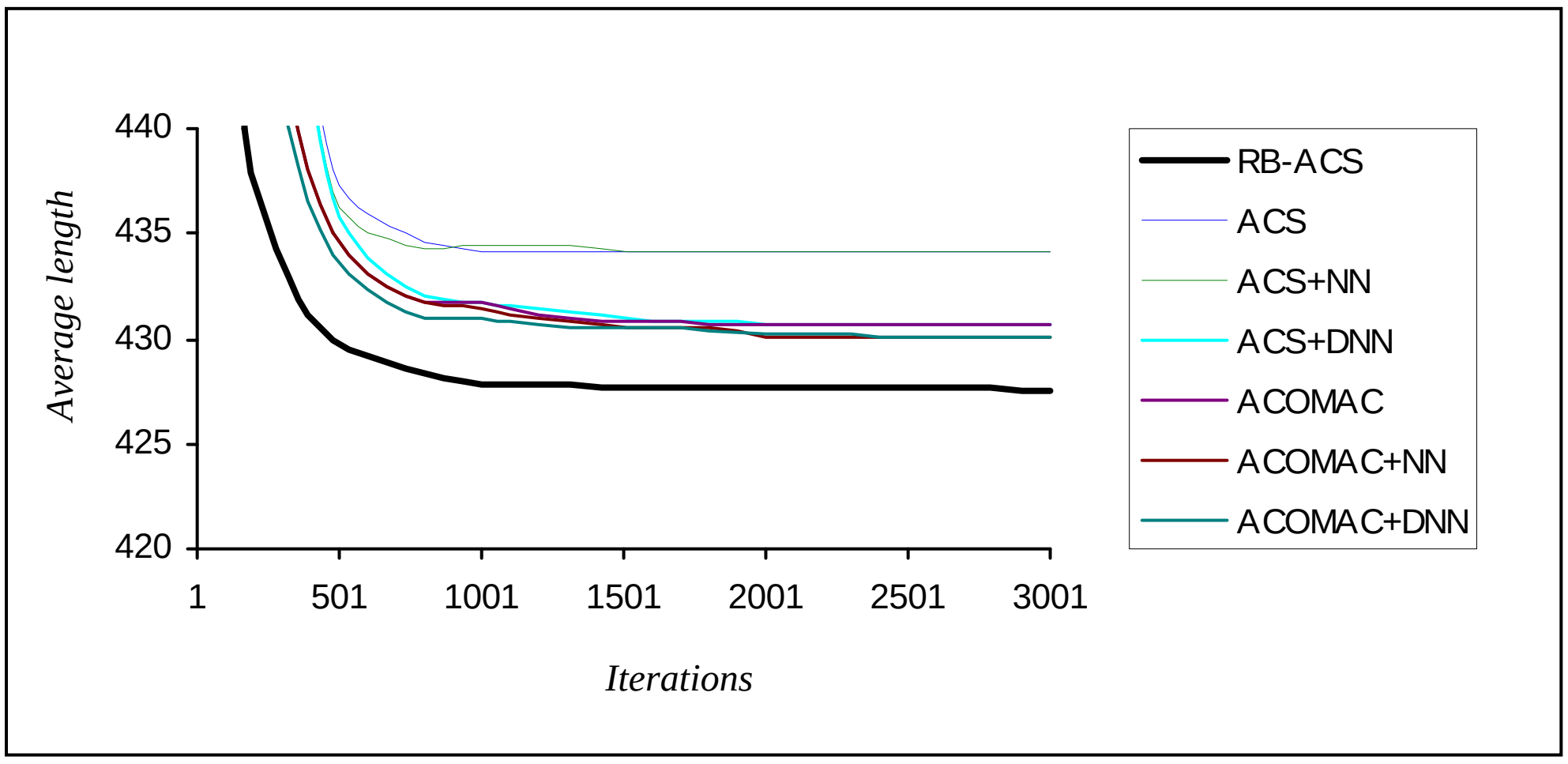

Fig. 3: Comparison of the proposed *RB-ACS* with ACS+DNN and ACOMAC+DNN using Eil51 benchmark problem with 51 cities

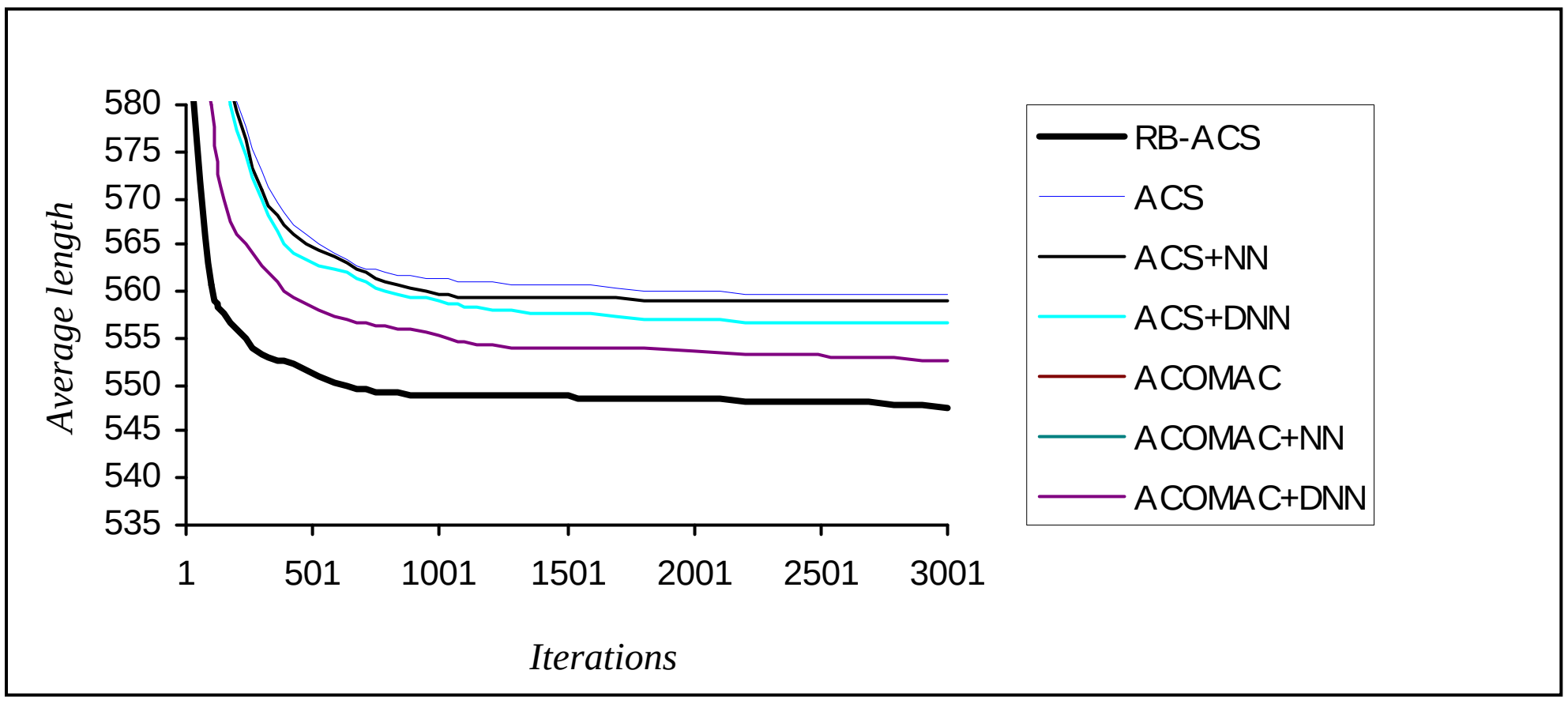

Fig. 4: Comparison of the proposed *RB-ACS* with ACS+DNN and ACOMAC+DNN using Eil76 benchmark problem with 76 cities

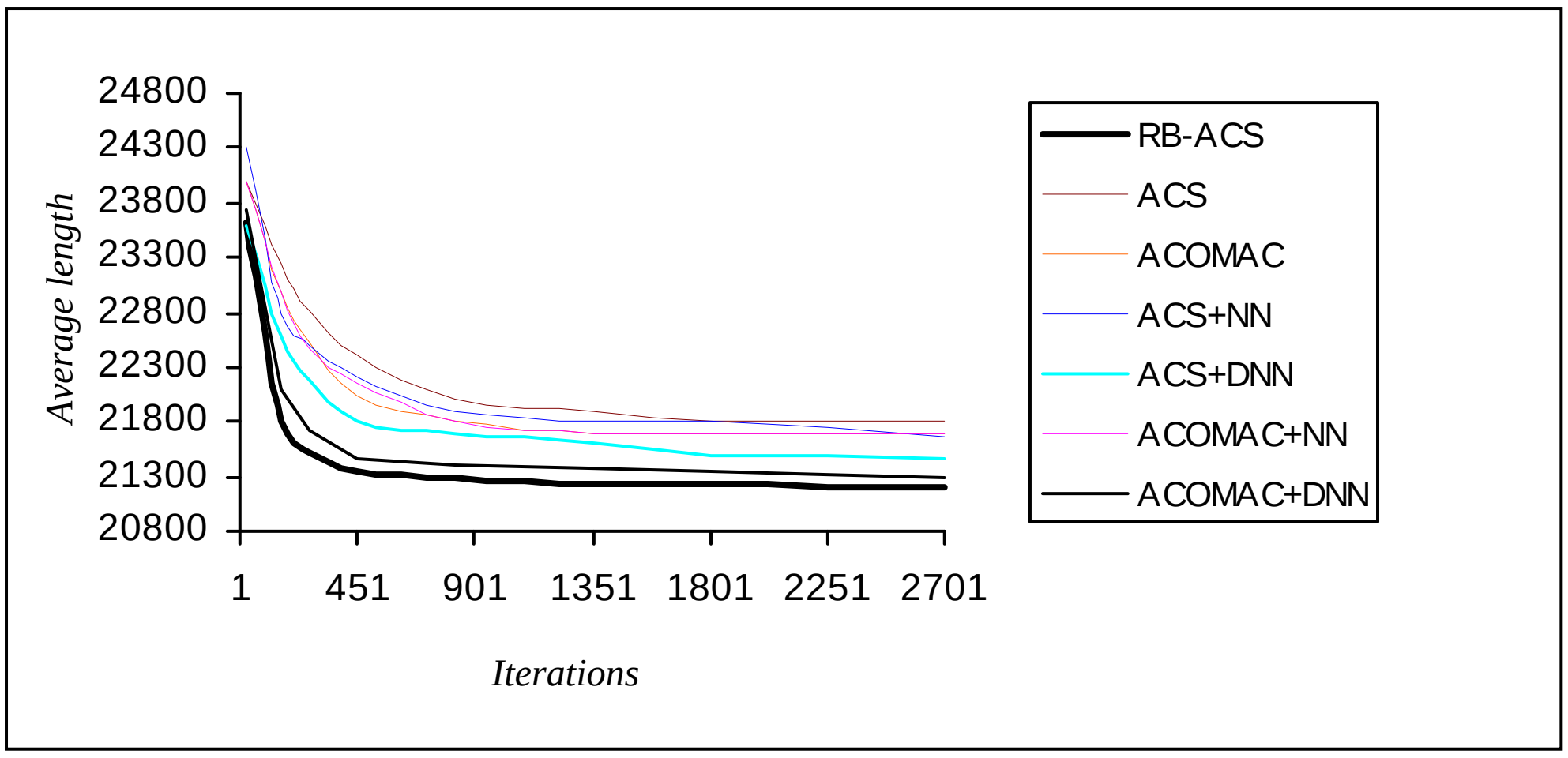

Fig. 5: Comparison of the proposed *RB-ACS* with ACS (include ACS+NN and ACS+DNN) and ACOMAC (include ACOMAC+NN and ACOMAC+DNN) using Kroa100 benchmark problem with 100 cities

**Table 1:** Comparison of *RB-ACS* with ACS (Including ACS+DNN) and ACOMAC (Including ACOMAC+DNN) [6] using four different TSP benchmark problems

| Benchmark problem(optimum) | ACS | ACOMAC | ACS+DNN | ACOMAC + DNN | RB-ACS |
|---|---|---|---|---|---|
| Eil51 (426) | 434.178 | 430.684 | 430.6656 | 430.0076 | 427.5 |
| Eil76 (538) | 559.7041 | 555.230 | 557.772 | 552.6128 | 549.333 |
| Kroa100 (21282) | 21684.64 | 21457.93 | 21440.0923 | 21408.23 | 21389.235 |

It is observed that the *RB-ACS* provides a significant improvement for obtaining a global optimum solution or near global optimum solution in a very short number of iterations.

## 5. CONCLUDING REMARKS

The proposed *Red-Black Ant Colony System* presented in this paper is shown to produce better results for the solution of larger traveling salesman problems. We believe and hope that our approach would be a very promising one because of its generality and because of its effectiveness in finding very good solutions efficiently to various fields of difficult problems. The fields of interest may be the difficult combinatorial problems such as load balancing problem in telecommunications networks, multiple-fuel economic load dispatch problem, binary constraint-satisfaction problem, random number generation, data mining, pattern recognition, job scheduling problem and much more. Actually, the *RB-ACS* is a promising approach which can be a valuable topic for the researchers from different fields, e.g., artificial intelligence, biology, mathematics operations research since it can greatly improve the performance for finding good solutions, especially for high-dimensional problems.

## REFERENCES

[1]. E.L. Lawler, J.K. Lenstra, A.H.G. Rinnooy-Kan and D.B. Shmoys, "The Travelling Salesman Problem," *New York:Wiley*, 1985.

[2]. J.L. Bentley, "Fast algorithms for geometric traveling salesman problems," *ORSA Journal on Computing*, Vol. 4, pp. 387–411, 1992.

[3]. M. Dorigo and L.M. Gambardella, "Ant Colony System: A Cooperative Learning Approach to the Traveling Salesman Problem," *IEEE Transactions on Evolutionary Computation*, Vol.1, No.1, pp. 53-66, 1997.

[4]. M. Dorigo, V. Maniezzo and A.Colorni, "The ant system: Optimization by a colony of cooperating agents," *IEEE Transactions on Systems, Man, and Cybernetics–Part B*, Vol. 26, No. 2, pp. 1-13, 1996.

[5]. T. Stützle and H. Hoos, "The ant system and local search for the traveling salesman problem," *Proceedings of ICEC'97 - 1997 IEEE 4th International Conference on Evolutionary Computation*, IEEE Press, 1997.

[6]. L.M. Gambardella and M. Dorigo, "Solving symmetric and asymmetric TSPs by ant colonies," *Proceedings of IEEE International Conference on Evolutionary Computation, IEEE-EC 96*, IEEE Press, pp. 622–627, 1996.

[7]. C. F. Tsai, C.W. Tsai and C.C. Tseng, "A new approach for solving large traveling salesman problem," *IEEE World Congress on Computational Intelligence*, [CD-ROM], 2002.

[8]. H. K. Tsai, J. M. Yang and C. Y. Kao, "Solving Traveling Salesman Problems by Combining Global and Local Search Mechanisms," *IEEE World Congress on Computational Intelligence*, [CD-ROM], 2002.

[9]. TSPLIB: http://www.iwr.uniheidelberg.de/iwr/ comopt/soft/TSPLIB95/TSPLIB.html.